\pdfoutput=1

\documentclass[11pt]{article}

\usepackage[]{emnlp2021}

\usepackage{times}
\usepackage{latexsym}

\usepackage{stmaryrd} 
\usepackage{linguex} 

\usepackage[T1]{fontenc}

\usepackage[utf8]{inputenc}

\usepackage{microtype}
\usepackage{multicol}

\usepackage{tikz}
\usetikzlibrary{positioning,arrows.meta,decorations.text,fit,shapes}

\newcommand{\T}[0]{\hspace*{0.6cm}} 
\renewcommand{\phi}[0]{\varphi} 

\usepackage{xparse}

\NewDocumentCommand\eval{sO{}O{}m}{%
  \IfBooleanTF#1
  {\ensuremath{\left\llbracket{#4}\right\rrbracket^{#2}_{#3}}}
  {\ensuremath{\left\llbracket\texttt{#4}\right\rrbracket^{#2}_{#3}}}
}

\NewDocumentCommand\free{sO{}O{}m}{%
  \IfBooleanTF#1
  {\ensuremath{\left\llbracket{#4}\right\rrbracket^{#2}_{#3}}}
  {\ensuremath{\mathbf{Free}\left(\texttt{#4}\right)^{#2}_{#3}}}
}

\title{Intensionalizing Abstract Meaning Representations:\\Non-Veridicality and Scope}

\author{Gregor Williamson \\
  Computer Science \\
  Emory University \\
  Atlanta, GA 30322, USA \\
  \resizebox{0.47\linewidth}{!}{\texttt{gregor.jude.williamson@emory.edu}} \\\And
  Patrick Elliott \\
  Linguistics and Philosophy \\
 Massachusetts Institute of Technology \\
Cambridge, MA 02139, USA \\
  \texttt{pdell@mit.edu} \AND
  {\bf Yuxin Ji}  \\
  Quantitative Theory and Methods \\
  Emory University \\
  Atlanta, GA 30322, USA \\
  \texttt{jessica.ji@emory.edu} \\ \And
  Jinho D. Choi\thanks{\hspace*{.1cm} Omitted from LAW-DMR'21 submission due to error.} \\
  Computer Science \\
  Emory University \\
  Atlanta, GA 30322, USA \\
  \texttt{jinho.choi@emory.edu}
}

\begin{document}
\maketitle


\begin{abstract}
    Abstract Meaning Representation (AMR) is a graphical meaning representation language designed to represent propositional information about argument structure.
    However, at present it is unable to satisfyingly represent non-veridical intensional contexts, often licensing inappropriate inferences.
    In this paper, we show how to resolve the problem of non-veridicality without appealing to layered graphs through a mapping from AMRs into Simply-Typed Lambda Calculus (STLC).
    At least for some cases, this requires the introduction of a new role \texttt{:content} which functions as an intensional operator.
    The translation proposed is inspired by the formal linguistics literature on the event semantics of attitude reports.
   Next, we address the interaction of quantifier scope and intensional operators in so-called de re/de dicto ambiguities.
    We adopt a scope node from the literature and provide an explicit multidimensional semantics utilizing Cooper storage which allows us to derive the de re and de dicto scope readings as well as intermediate scope readings which prove difficult for accounts without a scope node.
\end{abstract}
\section{Introduction}\label{sec:introduction}

Abstract Meaning Representation (AMR) is a graphical meaning representation in which graphs are rooted, directed, and acyclic \citep{banarescu-etal-2013-abstract}.
Non-terminal nodes are assigned variable IDs, terminal nodes are sense concepts (e.g., \texttt{believe-01}, \texttt{boy}, etc.) or constants (e.g., the polarity attribute \texttt{-}, cardinals, names, etc.), and labelled edges represent semantic relations between nodes.
The inventory of AMR disambiguated predicate senses (e.g., \texttt{believe-01}) are based on PropBank argument structure frames \citep{propbank,palmer-etal-2005-proposition, bonial-etal-2014-propbank}.
The graph in \ref{ex:graph} is an AMR for the sentence in \ref{ex:sentence}.
More commonly, however, AMRs are represented in Penman notation \citep{matthiessen1991text} as in \ref{ex:penman} or occasionally as a conjunction of logical triples \ref{ex:logic-trips}.

\ex. 
    \a. \textit{The boy hugged the dog.}\label{ex:sentence}\smallskip
    \b.
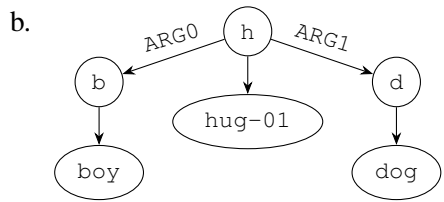
{\begin{tikzpicture}[baseline,auto,vertex/.style={draw,ellipse}]%
    \node[vertex] (h) {\small{\texttt{h}}};%
    \node[vertex,below left=0.2cm and 1.5cm of h] (b) {\small{\texttt{b}}};%
    \node[vertex,below=0.5cm of h] (hug) {\texttt{\small{hug-01}}};%
    \node[vertex,below=0.5cm of b] (boy) {\texttt{\small{boy}}};%
    \node[vertex,below right=0.2cm and 1.5cm of h] (d) {\small{\texttt{d}}};%
    \node[vertex,below=0.5cm of d] (dog) {\texttt{\small{dog}}};
    \path[-{Stealth[]}]%
      (h) edge [above left,pos=0.1] node [rotate=17] {\small{\texttt{ARG0}}} (b)%
      (b) edge 
      (boy)%
      (h) edge 
      (hug)%
      (d) edge 
      (dog)%
      (h) edge [above right,pos=0.1] node [rotate=-17] {\small{\texttt{ARG1}}} (d);%
  \end{tikzpicture}\medskip}\label{ex:graph}
    \c.{\texttt{(h\,/\,hug-01
    \\ \T :ARG0\,(b\,/\,boy)
    \\ \T :ARG1\,(d\,/\,dog))}
    \smallskip}\label{ex:penman}
    \d. {$\textsc{instance}(\texttt{h}, \texttt{hug-01})  
    \\ \T\wedge \textsc{instance}(\texttt{b}, \texttt{boy})  \wedge \texttt{ARG0}(\texttt{h}, \texttt{b}) 
    \\ \T \wedge \textsc{instance}(\texttt{d}, \texttt{dog}) \wedge \texttt{ARG1}(\texttt{h}, \texttt{d})$
    }\label{ex:logic-trips}

The main strength of AMR is its ability to represent argument structure, as these logical triples translate naturally into a rudimentary neo-Davidsonian event semantics \citep{Davidson1967,Parsons1990}, with every \textsc{instance} relation split into an existential quantifier and a one-place predicate.

\ex. {$\exists x(\mathbf{hug\textbf{-}01}(x) \wedge \exists y(\mathbf{boy}(y)  \wedge ARG0(x)(y) 
\\ \T \wedge \exists z(\mathbf{dog}(z) \wedge ARG1(x)(z))))$}

Recent developments have seen the expressive power of AMR improved both in terms of its graphic representation as well as its translation into logical forms.
For instance, AMR graphical representations have been enriched to represent Tense and Aspect \citep{donatelli-etal-2018-annotation,donatelli2019tense,van2021designing}, quantifier scope \citep{pustejovsky-etal-2019-modeling,van2021designing}, semantic number \citep{stabler2017}, and speech acts \citep{bonial-etal-2020-dialogue}, while translations into first and higher-order logics have been proposed as a means of capturing coreference \citep{artzi-etal-2015-broad} and quantifier scope \citep{bos-2016-squib,bos-2020-separating,stabler2017,lai-etal-2020-continuation}.

Despite these advances in theoretical work, AMR encounters issues when it comes to the semantics of intensional contexts.
For instance, \citet{crouch-kalouli-2018-named} note that an AMR like \ref{ex:bel-prop}, represented as a conjunction of logical triples, would permit an inference to \textit{`The girl is sick'} by conjunction elimination.

\ex.
\a.\textit{The boy believes that the girl is sick.}\label{ex:bel-sent}
\b.{\texttt{(b\,/\,believe-01
    \\ \T :ARG0\,(b2\,/\,boy)
    \\ \T :ARG1\,(s\,/\,sick-05
    \\ \T\T :ARG1\,(g\,/\,girl)))}\smallskip}\label{ex:bel-prop}

An even more striking consequence of this non-veridical problem is demonstrated by the following examples.\footnote{These examples are based on examples in the \textit{freakshow} section of the AMR guidelines, available at: \url{https://github.com/amrisi/amr-guidelines/blob/master/amr.md\#amr-freak-show}}

\ex. \label{ex:freakshow}
\a. \textit{The boy believes he is sick.}\label{ex:freakshow-example}
\b.\texttt{(b\,/\,believe-01
    \\ \T :ARG0\,(b2\,/\,boy)
    \\ \T :ARG1\,(s\,/\,sick-05
    \\ \T\T :ARG1\,b2))\smallskip}\label{ex:freakshow-example-b}
\c.\texttt{(b\,/\,believe-01
    \\ \T :ARG0\,(b2\,/\,boy
    \\ \T\T:ARG1-of\,(s\,/\,sick-05))
    \\ \T :ARG1\,s))}\label{ex:freakshow-example-c}

At first glance, it might appear that the predicate \texttt{sick-05} is in a non-veridical context in \ref{ex:freakshow-example-b} and a veridical context in \ref{ex:freakshow-example-c}.
This is because the grammatical subject position of the verb \textit{believe} is a veridical environment, while its sentential complement is not.
However, these graphs are in fact logically equivalent because for any AMR relation \texttt{R}, $\texttt{R}(x)(y) \Leftrightarrow \texttt{R-of}(y)(x)$. 
Consequently, \ref{ex:freakshow-example-b} and \ref{ex:freakshow-example-c} depict the same conjunction of logical triples once inverse relations are normalized.

\citet{crouch-kalouli-2018-named} provide a solution to the problem of non-veridicality in a graph-based representation by making use of a Graphic Knowledge Representation (GKR) \cite{kalouli2018gkr}, a layered graph in which identifiers, or names, are assigned to sub-graphs \citep{carroll2005named}.
A GKR separates conceptual structure (i.e., predicate-argument structure) and contextual structure into two sub-graphs.
The graph in \ref{ex:GKR-named} is a simplified GKR of \ref{ex:bel-sent}.
The lower context \texttt{bel} represents the intensional context of the boy's belief and is non-veridical (or averidical) with respect to the upper context \texttt{top}.
Consequently, the GKR does not permit the inference from \textit{'The boy believes the girl is sick'} to \textit{`The girl is sick.'}
 
\ex.
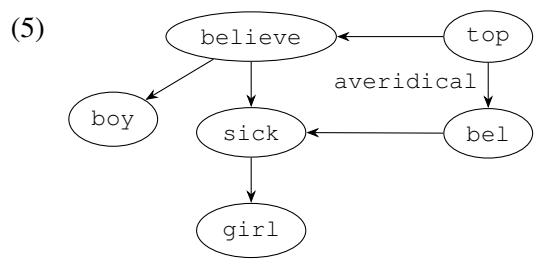
{\begin{tikzpicture}[baseline,auto,vertex/.style={draw,ellipse}]%
    \node[vertex] (believe) {\small{\texttt{believe}}};%
    \node[vertex,below left=.6cm and .6cm of believe] (boy) {\small{\texttt{boy}}};%
    \node[vertex,below=.6cm of believe] (sick) {\texttt{\small{sick}}};%
    \node[vertex,below=.6cm of sick] (girl) {\texttt{\small{girl}}};%
    \node[vertex,right=1.4cm of believe] (top) {\small{\texttt{top}}};%
    \node[vertex,below=.6cm of top] (bel) {\texttt{\small{bel}}};
    \path[-{Stealth[]}]%
      (believe) edge (boy)%
      (bel) edge (sick)%
      (sick) edge (girl)%
      (believe) edge (sick)%
      (top) edge [left, pos=0.4] node {\footnotesize{\texttt{averidical}}} (bel)%
      (top) edge (believe);%
  \end{tikzpicture}\smallskip}\label{ex:GKR-named}

Another problem for AMR's representation of attitude reports such as those containing the verb \textit{`believe'} is that PropBank argument structures often do not distinguish between propositional and non-propositional arguments.
For instance, the PropBank argument structure for \texttt{believe-01} assigns both nominal and clausal arguments the same argument role, \texttt{:ARG1}.\footnote{PropBank frames are available at: \url{https://github.com/propbank/propbank-frames/}}
Despite this, the PropBank frame includes a note that \textit{`believe'} can have both a theme and a propositional argument simultaneously (e.g., \textit{`Mary believed John that he didn't eat the last piece of pie'}).\footnote{The annotator concludes ``we could add an \texttt{:ARG2} [...] and use it only when \texttt{:ARG1} is already present, but that makes me sad, so let's just not mess with it until we actually see such an instance.''}

Given this lack of distinction between different types of arguments, it is not sufficient to lexically specify that the \texttt{:ARG1} of \texttt{believe-01} is always a non-veridical environment.
While this might work for \ref{ex:bel-prop}, it would also mean that we could not infer the existence of a girl from \ref{ex:bel-girl}.

\ex.\label{ex:bel-girl}
\a.\textit{The boy believed the girl.}
\b.{\texttt{(b\,/\,believe-01
    \\ \T :ARG0\,(b2\,/\,boy)
    \\ \T :ARG1\,(g\,/\,girl))}}

In what follows, we propose the introduction of an intensional relation \texttt{:content} responsible for introducing propositional arguments.
Replacing the \texttt{:ARG1} of \texttt{believe-01} in \ref{ex:freakshow} with the new \texttt{:content} role ensures firstly that \ref{ex:freakshow-example-b} contains a non-veridical environment and secondly that \ref{ex:freakshow-example-c} is not a representation for a coherent natural language sentence.
Crucially, the addition of \texttt{:content} and the translation function proposed for AMRs into logical forms offers a satisfying representation for intensional contexts without the need for additional graph structure.
Finally, we show how our logical forms interact with scope taking elements to derive attested interpretations of attitude reports with quantifier phrases (QPs).

\section{Extensional Semantics for AMRs}\label{sec:extensional-semantics}

We start by defining a simple translation for basic AMRs (without intensional operators or quantifiers) into the Simply-Typed Lambda Calculus (STLC).
Following \citet{bos-2016-squib} we define the syntax of AMRs recursively.
A simplex AMR is a constant \texttt{c}, variable \texttt{x}, or an instance assignment \texttt{(x\,/\,P)}.
Complex graphs are defined recursively as one or more subgraphs $\{\texttt{A}_1,\dots,\texttt{A}_n\}$ connected to an instance assignment by $n$ relations.

\ex. \texttt{A} $:=$
\a. \texttt{c}
\b. \texttt{x}
\c. \texttt{(x\,/\,P)}
\d. \texttt{(x\,/\,P\,:R$_1$\,A$_1$\ldots:R$_n$\,A$_n$)}

Our semantics for AMRs is a departure from that of \citet{bos-2016-squib} and \citet{lai-etal-2020-continuation}.
For now, we assume that the interpretation function $\llbracket.\rrbracket$ compositionally maps AMRs to a simple first-order calculus embedded in the STLC.%
\footnote{The STLC is widely used in analytical work on natural language semantics, and has a well-understood proof-theory and model-theory \citep{Carpenter1998}.} 
AMR constants and variables are mapped to STLC constants of type $e$, AMR predicates are mapped to STLC constants of type $e \rightarrow t$, and AMR roles are mapped to STLC constants of type $e \rightarrow e \rightarrow t$.\footnote{$e$ is the basic type of \textit{individuals}, and $t$ is the basic type of \textit{truth-values}. The constructor for functional types $a \rightarrow b$ is right-associative.}

\ex.
\a. $\eval{c} = \mathbf{c}$\hfill$e$
\b. $\eval{x} = x$\hfill$e$

Simple instance assignments are mapped to functional applications:

\ex. \textbf{Instance assignment}\\
$\llbracket \texttt{(x\,/\,P)} \rrbracket = P(x)$\hfill$t$

To state a semantics for complex AMRs, we first state a semantics for a \textit{role assignment} in \ref{def:role-assign}, consisting of a role and an embedded AMR, via pattern matching on the embedded AMR. A sequence of role assignments $\rho_1 \ldots \rho_n$ is interpreted via iterated conjunction \ref{def:role-seq}, and finally a complex AMR is interpreted by saturating a role sequence with the main variable of the AMR \ref{def:complex-amr}.

\ex.\label{def:role-assign} \textbf{Role assignment}
\a. $\eval{:R\,y} = \lambda x\,.\,R(x)(y)$
\b. $\eval{:R\,(y / $\ldots$ )}\\
= \lambda x\,.\,R(x)(y) \wedge \eval{(y / $\ldots$ )}$

\ex. \textbf{Role sequence}\\
$\eval*{\rho_1 \ldots \rho_n} = \lambda x\,.\,\eval*{\rho_1}(x) \wedge \ldots \wedge \eval*{\rho_n}(x)$\label{def:role-seq}

\ex. \textbf{Complex AMR}\\
$\eval{(x\,/\,P\,$\rho_1 \ldots \rho_n$)}\\
 = P(x) \wedge \eval*{\rho_1 \ldots \rho_n}(x)$\label{def:complex-amr}

Basic AMRs are thereby translated into simple conjunctive first-order formulae.

\ex. \textit{The boy admires himself.}
\a. $\texttt{(a\,/\,admire-01}\\
\T\texttt{:ARG0\,(b\,/\,boy)}\\
\T \texttt{:ARG1\,b))}$
\b. $\mathbf{admire\textbf{-}01}(a) \wedge ARG0(a)(b)\\
\T \wedge \mathbf{boy}(b) \wedge ARG1(a)(b)$

Finally, we declare an operation $\mathtt{close}$ (version 1) which, applied to a STLC expression $\phi$, introduces an existential quantifier which unselectively binds variables in the set $\mathbf{Free}(\phi)$.

\ex. \textbf{Close (version 1)}\\
$\mathtt{close}(\phi) = \exists x_1 \ldots x_n(\phi)$\\
\phantom{,}\hfill where $\{x_1,\ldots,x_n\} = \mathbf{Free}(\phi)$

Crucially for what follows, $\mathbf{Free}(\phi)$ does not correspond to the classical notion of free variables in $\phi$, but rather should be defined to ensure that recurrent variables are not bound by $\mathtt{close}$ prior to instance assignment. We define $\mathbf{Free}$ as follows:\footnote{We are grateful to an anonymous reviewer for pressing us on this point.}

\ex.
\a. $\free{c} = \emptyset$
\b. $\free{x} = \emptyset$
\c. $\free{(x\,/\,P)} = \{x\}$
\d. $\free{(x\,/\,P\,:R$_{1}$A$_{1}$\ldots R$_{n}$A$_{n}$)}$\\
$= \{x\} \cup \free{A$_1$} \cup \ldots \cup \free{A$_n$}$

Applying $\mathtt{close}$ to the example above returns an existential statement.

\ex. $\exists a,b.\mathbf{admire\textbf{-}01}(a) \wedge ARG0(a)(b)\\
\T \wedge \mathbf{boy}(b) \wedge ARG1(a)(b)$

Putting re-entrant nodes to one side, it is harmless to defer existential binding of free variables, since $\psi \wedge \exists x(\phi) \Leftrightarrow \exists x(\psi \wedge \phi)$ (if $x \notin \mathbf{Free}(\psi)$). This does away with the need to use continuation-passing style (c.f., \citealt{bos-2016-squib,lai-etal-2020-continuation}). Cases involving re-entrant nodes such as \textit{the dog scratched itself} are handled straightforwardly via matching variables.

\section{An Intensional Semantics}\label{sec:intensional-semantics}

Next, we systematically intensionalize the interpretation in a standard way by replacing the propositional type $t$ with $s \rightarrow t$ (the type of a function from \textit{worlds} to truth values; \citealt{Gallin1975}). Our existing interpretation procedure for basic AMRs remains largely intact, although we tweak the definitions of conjunction and existential quantification in order to accommodate the presence of additional world arguments.

\ex.
\a. $\phi \wedge_w \psi := \lambda w.\,\phi(w) \wedge \psi(w)$
\b. $\exists_w x(\phi) := \lambda w.\,\exists x(\phi(w))$

To provide a semantics for intensional operators such as attitude predicates in AMR, we adopt a variant of \citeauthor{kratzer2006decomposing}'s \citeyearpar{kratzer2006decomposing} Davidsonian event semantics for attitude verbs which has undergone a number of refinements \cite[e.g.,][]{moulton2009natural,elliott2016explaining,Elliott2020}.
More specifically, we propose a translation of AMRs of attitude reports into logical forms in which attitude events are associated with propositional content via a dedicated modal operator $\mathbf{cont}$.
In order to achieve this, we increase the AMR inventory of semantic roles with a \texttt{:content} role which is interpreted as a relation of type $e \rightarrow (s \rightarrow t) \rightarrow (s \rightarrow t)$.

\ex.
$\eval{:content\,A}\\
= \lambda x\,.\,\mathbf{cont}(x)(\mathtt{close}(\eval{A}))$

To see how this resolves the problem of non-veridicality consider again the two AMRs in \ref{ex:freakshow}, the first of which is repeated in \ref{ex:freak-derivation-1a} with the \texttt{:ARG1} role changed to \texttt{:content}.
The underlined argument shows that the world of evaluation for the translation of the \texttt{:content} argument is shifted to $w_2$, and thus we cannot infer that the boy is sick in $w$ as desired.

\ex. 
\a.{\texttt{(b\,/\,believe-01}
    \\ \T \texttt{:ARG0\,(b2\,/\,boy)}
    \\ \T \texttt{:content\,(s\,/\,sick-05}
    \\ \T\T \texttt{:ARG1\,b2))}
    \smallskip}\label{ex:freak-derivation-1a}
\b.{$\mathtt{close}(\llbracket \,\ref{ex:freak-derivation-1a}\, \rrbracket)$
    \\ $= \lambda w\, . \, \exists b,b_2(\mathbf{believe\textbf{-}01}(b)(w)
    \\ \T \wedge \mathbf{boy}(b_2)(w) 
    \\ \T \wedge ARG0(b)(b_2)(w) 
    \\ \T \wedge \, \mathbf{cont}(b)(\underline{\lambda w_2\,.\, \exists s(\mathbf{sick\textbf{-}01}(s)(w_2)}
    \\ \T\T \underline{\wedge ARG1(s)(b_2)(w_2))})(w))$}\label{ex:freak-derivation-1b}

Next, consider the same modification for \ref{ex:freakshow-example-c} repeated in \ref{ex:freak-derivation-2a}, modulo \texttt{:ARG1}$\Rightarrow$\texttt{:content}.

\ex.{\texttt{(b\,/\,believe-01
    \\ \T :ARG0\,(b2\,/\,boy
    \\ \T\T:ARG1-of\,(s\,/\,sick-05))
    \\ \T :content\,s))}
    }\label{ex:freak-derivation-2a}

The semantic rule for \texttt{:content} is only well-defined if the embedded AMR is an \textit{instance assignment}, due to the type of $\mathbf{cont}$. It follows that the interpretation of \ref{ex:freak-derivation-2a}, in which \texttt{:content} embeds a recurrent variable, is undefined. For this reason, AMRs in which \texttt{:content} embeds a constant or variable should be avoided by annotators.
The issue here is that reentrant nodes are used to model a diverse range of linguistic phenomena \citep{SzubertEtAl2020}, but are inappropriate for modelling anaphora to a proposition, such as \textit{`The boy who is sick believes it'}.%
\footnote{It should be noted that the AMR in \ref{ex:freakshow-example-c} in which \texttt{:ARG1} is not replaced with \texttt{:content} does have an interpretation which can be paraphrased as: \textit{`the boy who is in a state $s$ of being sick, believes the state $s$'}.
However, it is not clear that this corresponds to any coherent natural language sentence.}

\section{Content and Quantifier Scope: de re and de dicto Readings}

Another property of intensional operators is their ability to interact scopally with other operators such as quantifier phrases (QPs) (see \citealt{KeshetSchwarz2019} for a recent overview).
For example, the sentence in \ref{ex:de-re-de-dicto} could be paraphrased as in \ref{ex:de-re} which can be analyzed as an existential QP \textit{`a violin'} taking scope over the attitude predicate \textit{`hope'}, in which case the restrictor argument of the QP is evaluated in the actual world (i.e., there is an actual violin that the boy wants). In contrast, the reading in \ref{ex:de-dicto} can be analyzed as the existential being within the scope of the attitude (i.e., the boy hopes to be a violin-owner, but is not necessarily concerned about owning any particular violin).

\ex.\textit{The boy hopes to buy a violin.} \label{ex:de-re-de-dicto}
\a. There is a violin the boy hopes to buy.\\
\phantom{,}\hfill\textit{De re}\label{ex:de-re}
\b. The boy hopes to be a violin-owner.\\
\phantom{,}\hfill\textit{De dicto}\label{ex:de-dicto}

To capture these readings, we develop a semantics for AMRs enriched with additional graph structure for modelling scope \citep{pustejovsky-etal-2019-modeling} based on the mechanism of Cooper storage \citep{Cooper1983,kobele2018cooper}.

\subsection{Scope Semantics}\label{ssec:scope-semantics}

Before discussing the scope interaction of quantifiers and $\mathbf{cont}$, let us first develop a semantics for scope in non-intensional contexts.
Following \citet{pustejovsky-etal-2019-modeling} and \citet{van2021designing} we make use of a $\texttt{scope}$ node, which has a predicative argument representing the core argument structure and reentrant variables to represent the order of quantifier scope.
For example, in \ref{ex:scope-node}, the QP of $\texttt{:ARG0}$ scopes over that of $\texttt{:ARG1}$.

\ex. 
\a. A computer is on every desk.
\b. $\texttt{(s\,/\,scope}\\
\T\texttt{:pred(b\,/\,be-located-at-91}\\
\T\T\texttt{:ARG0(c\,/\,computer)}\\
\T\T\texttt{:ARG1(d\,/\,desk}\\
\T\T\T\texttt{:quant\,every))}\\
\T\texttt{:ARG0\,d}\\
\T\texttt{:ARG1\,c)}$\label{ex:scope-node}

Next, we define an explicit interpretation of scope nodes using Cooper storage \citep{Cooper1983,kobele2018cooper}.
A \textit{store} is an \textit{assignment} of variables to STLC expressions of type $(e \rightarrow t) \rightarrow t$.\footnote{We leave world variables implicit in section \ref{ssec:scope-semantics} for the sake of readability.}

\ex. $\{(x,\mathbf{every}(\mathbf{boy})),(y,\mathbf{some}(\mathbf{girl})\}$

Instead of simply mapping AMRs to expressions of STLC, we map them to a pair consisting of a store $s$ and an ordinary semantic value (i.e., an STLC expression).
We assume the following notational conventions.

\ex.
\a. $\eval{A} = s \cdot \phi$
\b. $\eval[][s]{A} = s$
\c. $\eval[][o]{A} = \phi$

In this system, AMRs like those we have considered so far update the store vacuously \ref{ex:vacuous-update} while retaining their ordinary semantic value \ref{ex:ordinary-value}.

\ex. \textbf{Instance assignment (revised)}
\a. $\eval[][s]{(d\,/\,dance-01)} = \emptyset$\label{ex:vacuous-update}
\b. $\eval[][o]{(d\,/\,dance-01)} = \mathbf{dance\textbf{-}01}(d)$\label{ex:ordinary-value}

In order to illustrate how \texttt{:quant} is interpreted, we begin by stating a semantics for an instance assignment decorated with a single \texttt{:quant} role.\footnote{We generalize this rule to the more complex case involving \texttt{:quant} and a sequence of role assignments, as well as cases of nested quantification in appendix \ref{sec:appendix-nested}.}
We assume that AMR determiner constants such as \texttt{every} are mapped to STLC constants of type $(e \rightarrow t) \rightarrow (e \rightarrow t) \rightarrow t$, such as the quantificational determiner $\mathbf{every}$ \citep[e.g.,][]{BarwiseCooper1981}.
The store is updated with a generalized quantifier constructed from the determiner and the property in the instance assignment \ref{def:quant-update}.
The ordinary semantic value, on the other hand, is the truth constant $\top$, the addition of which is redundant in a string of conjunctions \ref{def:ord-update}.

\ex. \textbf{Quantifier storage}
\a. $\eval[][s]{(x\,/\,P\,:quant\,D)}\\= \{(x,\mathbf{D}(P))\}$\label{def:quant-update}
\b. $\eval[][o]{(x\,/\,P\,:quant\,D)}= \top$\label{def:ord-update}

On this semantics, basic AMRs have a non-trivial ordinary value but perform a vacuous store update.
Conversely, QPs perform a non-trivial store update but have a redundant ordinary semantic value. Crucially, however, QPs still contribute one or more variables to the ordinary value of a role assignment via pattern matching, as in \ref{def:role-assign-rev} below.

We revise our translation function to ensure that the store gets passed up during the derivation.

\ex. \textbf{Role assignment (revised)}\\
$\eval{:R\,(y\,/\,$\ldots$)}\\
= \eval[][s]{(y\,/\,$\ldots$)}\\
\T\cdot \lambda x\,.\,R(x)(y) \wedge \eval[][o]{(y\,/\,$\ldots$)}$\label{def:role-assign-rev}

\ex. \textbf{Role sequences (revised)}\\
$\eval*{\rho_1 \ldots \rho_n}\\
= \eval*[][s]{\rho_1} \cup \ldots \cup \eval*[][s]{\rho_n}\\
\T\cdot \lambda x\,.\,\eval*[][o]{\rho_1}(x) \wedge \ldots \wedge \eval*[][o]{\rho_n}(x)$

\ex. \textbf{Complex AMR (revised)}\\
$\eval{(x\,/\,P\,$\rho_1 \ldots \rho_n$)}\\
= \eval*[][s]{\rho_1\ldots\rho_n}\\
\T\cdot P(x) \wedge \eval*[][o]{\rho_1\ldots\rho_n}(x)$

In combination with \ref{def:quant-update}, this ensures that the store of a complex AMR will contain the index-quantifier pairs added to the store by its subgraphs, as shown in the following example.

\ex. \textit{Every boy danced.} 
\a.{$\texttt{(d\,/\,dance}\\
\T \texttt{:ARG0\,(b\,/\,boy} \\
\T\T \texttt{:quant\,every))}$\smallskip}\label{ex:every-boy-dance}
\b.{$\llbracket\,\ref{ex:every-boy-dance}\,\rrbracket = \\ 
\T\{(b,\mathbf{every}(\mathbf{boy}))\}\\
\T \T\cdot \mathbf{dance}(d) \wedge ARG0(d)(b)$}

In order to retrieve the quantifier from the store, we declare an operation $\mathtt{pop}_x$ which, given a variable $x$, store $s$, and logical form $\phi$ retrieves the expression in $s$ paired with $x$, and applies it to $\lambda x\,.\,\phi$.
We write $s_x$ for the expression in $s$ paired with $x$.%

\ex. \textbf{Pop} \\
$\mathtt{pop}_x(s,\phi) = s - \{(x, s_x)\} \cdot s_x(\lambda x\,.\,\phi)$

We also restate our $\mathtt{close}$ operation which now existentially binds any free variables which are not associated with an index in the store.

\ex. \textbf{Close (version 2)}\\
$\mathtt{close}(s,\phi) = s \cdot \exists x_1 \ldots x_n(\phi)\\
\T\T\{x_1,\ldots ,x_n\}\\
\T\T= \mathbf{Free}(\phi) - \{v\,|\,(v,*) \in s\}$
  
As mentioned above, scope nodes are decorated with roles embedding reentrant nodes indicating scope-takers (\texttt{:ARG$n$}), and a role to indicate the scope site (\texttt{:pred}).
We state the semantics for a complex AMR headed by a scope node syncategorematically: a scope node with arguments $x_1$,\ldots,$x_n$ induces evaluation of the quantifiers stored at $x_1$,\ldots,$x_n$.%
\footnote{\texttt{x}$_n$ is associated with \texttt{:ARG$n_{-1}$} because the indexing of \texttt{:ARG}s starts at zero.}%
$^{,}$\footnote{Note that this property of scope nodes ensures that event quantification takes narrow scope with respect to other operators in the sentence \citep{champollion2015interaction}.}

\ex. \textbf{Interpreting scope nodes\smallskip}\\
$\eval*{
\begin{array}{@{\,}l@{\,}}
\texttt{(s\,/\,scope}\\
\T\texttt{:ARG0\,x$_1$\,\ldots\,:ARG$n_{-1}$\,x$_n$}\\
\T\texttt{:pred\,A)}
\end{array}
}\\
= \mathtt{pop}(x_1)\\
\T(\ldots(\mathtt{pop}(x_n)\\
\T\T(\mathtt{close}(\eval{A})))\ldots)$

Consider again the example \textit{`every boy danced'}, but now with a scope node.

\ex.
\a.\texttt{(s\,/\,scope
\\\T:ARG0\,b
\\\T:pred\,(d\,/\,dance-01
\\\T\T:ARG0\,(b\,/\,boy\,
\\\T\T\T:quant\,every)))}\label{ex:every-boy-dance-scopenode}
\b.{$\llbracket\,\ref{ex:every-boy-dance-scopenode}\,\rrbracket = \\ \T \emptyset \cdot \mathbf{every}(\mathbf{boy})
\\ \T\T (\lambda b\,.\,\exists d(\mathbf{dance\textbf{-}01}(d)
\\ \T\T\T \wedge ARG0(d)(b)))$}

\subsection{Deriving the de re and de dicto readings}\label{ssec:deriving-de-re-de-dicto}

Now we reintroduce world variables to see how this interpretation function can translate both the de re and de dicto readings of \ref{ex:de-re-de-dicto} above, starting with the de re reading.
In \ref{ex:de-re-derivation}, the variable \texttt{v} is the \texttt{:ARG0} of the scope node and consequently the QP \textit{a violin} takes scope over the attitude verb.

\ex.
\a.{\texttt{(s\,/\,scope \\
\T:ARG0\,v\\
\T:pred\,(h\,/\,hope-01\\
\T\T:ARG0\,(b\,/\,boy)\\
\T\T:content\,(b2\,/\,buy-01
\T\T\T:ARG0\,b
\T\T\T:ARG1\,(v\,/\,violin
\T\T\T\T:quant a)))))}\smallskip}\label{ex:de-re-derivation}
\b.{$\llbracket\, \ref{ex:de-re-derivation}\, \rrbracket = \emptyset \cdot\\
 \lambda w \,.\, \exists v,b,h(\mathbf{violin}(v)(w)\\
\T \wedge \mathbf{hope\textbf{-}01}(h)(w) \\
\T \wedge \mathbf{boy}(b)(w)\\
\T \wedge ARG0(h)(b)(w) \\
\T \wedge \mathbf{cont}(h)\\
\T\T(\lambda w_2\,.\, \exists b_2(\mathbf{buy\textbf{-}01}(b_2)(w_2)\\
\T\T\T \wedge ARG0(b_2)(b)(w_2)\\
\T\T\T \wedge ARG1(b_2)(v)(w_2)))(w))$}

Here, there is a specific violin in the world of evaluation $w$ which the boy hopes to buy.

For the de dicto reading, we do not need to do anything special.
However, to close off the interpretation we can either embed the entire AMR under a scope node or use the $\mathbf{close}$ operation to bind any free variables.

\ex.
\a.{\texttt{(h\,/\,hope-01\\
\T:ARG0\,(b\,/\,boy)\\
\T:content\,(b2\,/\,buy-01
\T\T:ARG0\,b
\T\T:ARG1\,(v\,/\,violin))))}\smallskip}\label{ex:de-dicto-derivation}
\b.{$\mathbf{close}(\llbracket \ref{ex:de-dicto-derivation} \rrbracket) = \emptyset \cdot \\
\lambda w\,.\, \exists h,b(\mathbf{hope\textbf{-}01}(h)(w) \\
\T \wedge \mathbf{boy}(b)(w)\\
\T \wedge ARG0(h)(b)(w)\\
\T \wedge \mathbf{cont}(h)\\
\T\T(\lambda w_2\,.\, \exists b_2,v (\mathbf{buy\textbf{-}01}(b_2)(w_2) \\
\T\T\T \wedge \mathbf{violin}(v)(w_2) \\
\T\T\T \wedge ARG0(b_2)(b)(w_2)\\
\T\T\T \wedge ARG1(b_2)(v)(w_2)) )(w))$}

Here, the existential is within the scope of the lambda operator $\lambda w_2$ and the restrictor argument $\mathbf{violin}$ is evaluated in the boy's hope worlds $w_2$.

\subsection{Intermediate de dicto reading}\label{ssec:intermediate-de-dicto}

Although \citet{pustejovsky-etal-2019-modeling} frame their proposal as ``embed[ding an AMR] under a scope graph'', our implementation also permits the embedding of a scope node within an AMR.
Doing so allows us to derive intermediate scope readings.
Consider the following example.

\ex.\textit{The boy thinks the girl hopes to buy a violin.}
\a. The boy thinks there is a violin that the girl hopes to buy.

In this intermediate reading, the intensional object might not be a violin in the actual world, nor in the girl's desire worlds, rather it is a violin in the boys belief worlds.
\citet{stabler2017} notes that accounts such as \citet{bos-2016-squib}, and later \citet{lai-etal-2020-continuation}, cannot capture these sorts of intermediate scope readings since their projective semantics always derives widest scope of QPs.
However, this reading can be captured on the present account straightforwardly from the following structure.%
\footnote{\citet{van2021designing} adopt the scope node approach in combination with a variant of \citeauthor{lai-etal-2020-continuation}'s semantics. Such an approach should also derive intermediate scope readings.}

\ex.{\texttt{(t\,/\,think-01\\
\T:ARG0\,(b\,/\,boy)\\
\T:content\,(s\,/\,scope\\
\T\T:ARG0\,v\\
\T\T:pred\,(h\,/\,hope-01\\
\T\T\T:ARG0\,(g\,/\,girl)\\
\T\T\T:content\,(b\,/\,buy-01\\
\T\T\T\T:ARG0\,g\\
\T\T\T\T:ARG1\,(v\,/\,violin
\T\T\T\T\T:quant a)))))%
}}

\subsection{Problematic Scope}\label{ssec:non-specific-de-re}

Besides the de re and de dicto readings, \citet{fodor1970} considers two further readings in which the restrictor argument of the QP is interpreted separately from the scopal position of the quantifier.

\ex.\textit{The boy hopes to buy a violin.}\label{ex:non-spec-de-re-spec-de-dicto}
\a. {There are things which are violins and the boy hopes to buy one of them. \\ \hspace*{\fill} \textit{Non-specific de re}}\label{ex:non-spec-de-re}
\b. \# There is a thing the boy hopes to buy which he believes is a violin. \\ \hspace*{\fill}\textit{Specific de dicto}\label{ex:spec-de-dicto}

Although there is a general consensus in the literature that \ref{ex:spec-de-dicto} is not a possible reading of \ref{ex:non-spec-de-re-spec-de-dicto}, the reading in \ref{ex:non-spec-de-re} is possible.
At present, we cannot derive the reading in \ref{ex:non-spec-de-re}.
While it is possible to enrich AMRs further to accommodate such interpretations, it remains to be seen whether such an effort is worthwhile.
Although such interpretations are attested and theoretically significant, they are not common, and accounting for such interpretations would likely involve enriching the graphical representation of AMRs in ways which would make them far less tractable for annotators and parsers.
We leave it to future research to determine whether we can accommodate this third reading without unintentionally complicating the AMRs in undesirable ways.

\section{Discussion}\label{sec:discussion}

We have proposed to extend the expressive power of AMR in two respects.
Firstly, we have enriched the graphical form of AMRs by increasing the inventory of AMR roles with the role \texttt{:content}.
We did so in order to distinguish between intensional and non-intensional arguments of modal operators such as attitude predicates.
Secondly, we provided a translation of AMRs into logical forms which allowed us to solve the problem of non-veridicality as well as complex scope interactions between quantifiers and intenstional operators.

The addition of \texttt{:content} has ramifications for AMR annotation as well as backwards compatibility.
We opted to avoid proposing the addition of new numbered argument roles for predicate like \texttt{believe-01} since this would involve a wide-scale revision of AMR's frameset as well as complicating any effort to convert between existing corpora and an enriched corpus, as this would involve a many-to-many mapping.
Instead, we proposed a semantically motivated intensional relation \texttt{:content} which will reduce the complexity of any conversion effort, requiring only a many-to-one mapping.

When designing meaning representations, there is inevitably a trade-off between how adequate the representation is (e.g., how much semantic information is present) and how tractable the representations are for large scale annotation projects.
Although we believe that \texttt{:content} should not be any more difficult to annotate than non-core roles in AMR, it is likely that annotators may struggle to resolve and represent quantifier scope and de re/de dicto ambiguities.
Thankfully, in addition to being representationally adequate, the scope node approach adopted from \citet{pustejovsky-etal-2019-modeling} and \citet{van2021designing} is both intuitive and transparent.
Future research could aim to gauge the level inter-annotator agreement when representing such phenomena in a small scale annotation task.

Beyond AMR, the enriched graphical language Uniform Meaning Representation (UMR) \citep{van2021designing} also utilizes scope nodes to capture quantifier scope relations.
The \texttt{:content} role and translation function proposed here can be adopted wholesale for UMR, and thus will prove useful in future annotation projects for this more expressive annotation scheme.

Finally, the \texttt{:content} role and its intensional translation may also facilitate downstream NLP tasks. Specifically, different attitude predicates trigger different lexical inferences regarding the truth of their complement depending on whether they are factive (e.g., \texttt{know-01}), counterfactive (e.g., \texttt{pretend-01}) or non-veridical (e.g., \texttt{believe-01}).
A number of rich resources exist in this domain including the MegaVeridicality data sets \citep{white2018role, white-etal-2018-lexicosyntactic} which contain factuality judgments for a comprehensive list of English verbs that embed finite clauses as well as a variety of predicates which embed non-finite clauses.
Resources such as these may be deployed alongside AMR for NLI, since the logical forms which our interpretation function produces represent the scope of the attitude verb, unlike in a flat list of logical triples.

A python script for translating AMR into STLC is available at \url{https://github.com/emorynlp/Intensionalizing-AMR}, and we are currently working on developing a script to convert intensional uses of numbered arguments into the new \texttt{:content} role which we also plan to make publicly available.
\section{Conclusion}\label{sec:conclusion}

Abstract Meaning Representations (AMRs) are unable to represent non-veridical environments in a semantically satisfying way.
When an AMR is translated into a conjunctions of logical triples, it permits spurious inferences to the truth of any of its subgraphs via conjunction elimination.
We proposed to rectify this through the introduction of a novel AMR role \texttt{:content}, before providing an intensional interpretation for AMRs which correctly invalidates such inferences.
We then showed how such a semantics can be combined with a means of modelling scope to derive de re and de dicto readings of natural language sentences with intensional operators and quantifiers.
We concluded that the inclusion of a scope node \citep{pustejovsky-etal-2019-modeling,van2021designing} is necessary in order to capture intermediate scope readings, and we provided a translation function from AMRs into STLC permitting the derivation of complex interactions of natural language quantifiers with intensional operators.
This work is part of the concerted effort to increase the expressive power of AMRs while also maintaining tractable representations ensuring that large scale annotation projects can be performed with minimal instruction.

\bibliography{main}
\bibliographystyle{acl_natbib}

\appendix

\section{Cooper storage and nested quantification}\label{sec:appendix-nested}

In the main body, we state a semantics for instance assignments decorated with \texttt{:quant}. Here, we state a more general rule for interpreting an instance assignment decorated with \texttt{:quant} and an additional sequence of role assignments $\rho_1 \ldots \rho_2$. This is necessary for interpreting AMRs arising from sentences involving nested quantifiers, such as \ref{ex:nested}, which may correpond to a \texttt{scope}-enriched AMR such as \ref{amr:nested}, in which a \texttt{:quant} role co-occurs with an embedded AMR, itself decorated with a \texttt{:quant} role.

\ex. 
\a. \textit{Every class with a certain two professors is difficult.}\label{ex:nested}
\b. \texttt{(s\,/\,scope\\
\T:ARG0\,p\\
\T:ARG1\,c}\\
\T\texttt{:pred\,(d\,/\,difficult}\\
\T\T\texttt{:domain\,(c\,/\,class}\\
\T\T\T\texttt{:quant\,every}\\
\T\T\T\texttt{:prep-with\,(p\,/\,prof.\\
\T\T\T\T:quant\,2))))}\label{amr:nested}

The rule for interpreting complex AMRs with a \texttt{:quant} role and an additional role sequence is given below. The idea is that stores associated with embedded AMRs are passed up, and the determiner introduce by \texttt{:quant} takes scope over the role sequence.

\ex.
$\eval[][s]{(x\,/\,P\,:quant\,D\,$\rho_1\ldots\rho_n$)}\\ = 
\eval*[][s]{\rho_1\ldots \rho_n}\\
\cup \{(x,\mathbf{D}(\lambda x\,.\,P(x) \wedge \eval*[][o]{\rho_1\ldots\rho_n}(x))\}$

\ex.
$\eval[][o]{(x\,/\,P\,:quant\,D\,$\rho_1\ldots\rho_n$)} = \top$

This more sophisticated interpretation rule allows us to interpret AMRs such as that in \ref{amr:nested}. The derivation involves storing a quantificational expression containing a free variable, which comes to be bound once the scope node is resolved. In \ref{nested:partial} we provide the interpretation of the AMR embedded under \texttt{:pred} in \ref{amr:nested}. A full derivation is left as an exercise to the reader.

\ex.\label{nested:partial}\hspace*{-.25cm}
$\left\{\begin{array}{@{}c@{}}
     (p,\mathbf{two}(\mathbf{professor}),  \\
 (c,\mathbf{every}(\lambda x.\,\mathbf{class}(x)\wedge \mathbf{with}(x)(p)))
\end{array}\\
\right\}\\
\T\cdot \mathbf{difficult}(d) \wedge \mathbf{domain}(d)(c)$

As a final note, care has to be taken to ensure that the quantifiers are evaluated in a certain order in cases involving nested quantification. Concretely, if the values of \texttt{:ARG0} and \texttt{:ARG1} are flipped, in the AMR in \ref{amr:nested}, then the free variable in the stored universal quantifier will remain free post-evaluation, which does not correspond to an attested interpretation of \ref{ex:nested}. Besides placing an implausible cognitive load on annotators, this is a known deficiency of the rudimentary Cooper storage system adopted here \citep{Keller1988}. More sophisticated approaches to Cooper storage which avoid this issue \citep{kobele2018cooper} could be adapted for the purpose of interpreting AMRs; we leave this to future work.

\end{document}